\documentclass[11pt,letterpaper]{article}
\usepackage[T1]{fontenc}
\usepackage[utf8]{inputenc}
\usepackage{lmodern}
\usepackage[margin=1in]{geometry}
\usepackage{microtype}
\usepackage{amsmath,amssymb}
\usepackage{graphicx}
\usepackage{array,booktabs,tabularx}
\usepackage{float}
\usepackage[section]{placeins}
\usepackage{needspace}
\usepackage{caption}
\usepackage{enumitem}
\usepackage{listings}
\usepackage[numbers,sort&compress]{natbib}
\usepackage{xurl}
\usepackage[hidelinks]{hyperref}

\newcolumntype{L}{>{\raggedright\arraybackslash}X}
\newcolumntype{P}[1]{>{\raggedright\arraybackslash}p{#1}}

\newcommand{\CNYmu}{CNY\,mu$^{-1}$}

\newcommand{\declaration}[2]{\par\addvspace{6pt}\noindent\textbf{#1} #2\par}
\title{Integrating Domain Adaptation and Retrieval-Augmented Generation for Saline-Alkali Land Reclamation Planning}
\author{%
  \normalsize Zuguang~He, Congcong~Zheng, Wanyue~Chen, Wenya~Zhou, Zaijin~Yang,\\
  \normalsize Long~Long, Yucheng~Zhang and Shuqin~Gao\thanks{Corresponding author: \href{mailto:gaoshuqin@ict.ac.cn}{gaoshuqin@ict.ac.cn}.}\\[0.4em]
  \small Institute of Computing Technology, Chinese Academy of Sciences\\
  \small Beijing 100190, China\\[0.3em]
  \scriptsize \texttt{hezuguang@ict.ac.cn}, \texttt{zhengcongcong@ict.ac.cn}, \texttt{chenwanyue@ict.ac.cn}\\
  \scriptsize \texttt{zhouwenya@ict.ac.cn}, \texttt{yangzaijin@ict.ac.cn}, \texttt{longlong@ict.ac.cn}, \texttt{zhangyucheng@ict.ac.cn}
}
\date{}

\newcommand{\ContributionStatement}{The order of authorship reflects the authors' relative contributions to this work. Conceptualization, S.G., Z.H. and C.Z.; software, Z.H.; investigation, Z.H. and C.Z.; validation, Z.H., C.Z. and W.C.; formal analysis, Z.H.; writing---original draft preparation, Z.H.; writing---review and editing, Z.H. and C.Z.; funding acquisition, S.G., Z.Y., W.Z., L.L. and Y.Z.}
\newcommand{\FundingStatement}{Supported by the Chinese Academy of Sciences.}

\newcommand{\EthicsStatement}{Institutional permission was obtained from the Institute of Computing Technology, Chinese Academy of Sciences, for the non-interventional expert-questionnaire assessment. Expert participation was voluntary and based on informed consent, and evaluation results are presented in aggregate form. The study involved literature-based computational analyses and professional assessment of model-generated reclamation plans; no clinical interventions or animal experiments were conducted.}
\newcommand{\ConflictStatement}{The authors declare no conflicts of interest.}
\newcommand{\ConsentStatement}{Informed consent was obtained from all experts involved in the evaluation.}
\newcommand{\DataAvailabilityStatement}{The derived SA-Bench data, implementation code, and evaluation questionnaires supporting this study will be made publicly available in a future release. Released expert-evaluation records will be anonymized, and access to the original scientific papers remains subject to the respective publishers' licenses and database terms.}

\hypersetup{
  pdftitle={Integrating Domain Adaptation and Retrieval-Augmented Generation for Saline-Alkali Land Reclamation Planning},
  pdfauthor={Zuguang He, Congcong Zheng, Wanyue Chen, Wenya Zhou, Zaijin Yang, Long Long, Yucheng Zhang, Shuqin Gao},
  pdfsubject={Saline-alkali land reclamation planning},
  pdfkeywords={salt-affected soils, agricultural decision support, large language model, retrieval-augmented generation, domain adaptation}
}

\begin{document}
\maketitle
\begin{abstract}
Reclamation of saline-alkali land requires management plans that account for soil and water conditions, treatment evidence, and financial constraints. This study develops Wenyao, an agricultural decision-support system integrating domain adaptation with retrieval-augmented generation. A bilingual corpus of 8,664 research papers was processed using a 110-indicator schema to construct SA-Bench, comprising 4,860 paper-level planning instances. A QwQ-32B model was adapted through low-rank supervised fine-tuning and combined with case retrieval for budget-conditioned planning. Evidence extraction achieved an F1-score of 0.83 on 300 expert-annotated papers. Without retrieval, Wenyao achieved 95.83\% accuracy on 936 source-disjoint multiple-choice questions, compared with 87.93\% for the strongest general-purpose baseline; removing explanatory supervision reduced accuracy to 89.42\%. Matched-evidence comparisons across 69 region--budget cases supported complementary contributions from adaptation and retrieval. In a blind assessment by 31 experts, Wenyao with retrieval obtained the highest mean scores for objectives and projected outcomes, whereas GPT-5 scored highest for cost estimates. These findings support literature-informed reclamation planning with professional review, while ongoing field monitoring will assess soil and crop responses.
\end{abstract}


\section{Introduction}

Salt-affected soils constrain agricultural production through interacting effects on crop water uptake, ion balance, soil structure, and nutrient availability~\cite{munns2008mechanisms,rengasamy2010processes}. Reclamation consequently requires more than lowering a single soil indicator: management must address the processes responsible for salinity or sodicity and support crop establishment under local water and resource constraints. The environmental processes underlying salinization vary across landscapes~\cite{rengasamy2006world}. Global analyses describe spatial variation in historical salinity and sodicity and project changes in primary salinization under future climates~\cite{hassani2020predicting,hassani2021global}. Reviews of reclamation in China emphasize the need to integrate soil improvement with productive land use~\cite{wang2024saline,zhu2024health}. Selecting an appropriate combination of measures for a particular field remains an important agricultural decision problem.

The suitability of an intervention depends on the conditions under which it will be implemented. Fields with similar pH may differ in electrical conductivity (EC), exchangeable sodium percentage (ESP), texture, groundwater depth, drainage, and irrigation-water quality. Reviews of salt-affected soil management describe engineering, chemical, agronomic, biological, and combined approaches, with no single treatment suitable for all sites~\cite{shaygan2022reclamation,toth2022review}. For example, evidence from combined organic-amendment and gypsum treatments must be interpreted in relation to the tested soil and experimental conditions~\cite{foronda2022gypsum}. Irrigation-water salinity and water management further constrain the transfer of recommendations between locations~\cite{qadir2004management,tessema2023salinity}. A useful planning tool must therefore relate initial conditions to treatment mechanisms, implementation requirements, and the responses actually reported in comparable studies.

Recent amendment research further illustrates the need to retain treatment context. A meta-analysis of biochar applications relates responses to environmental conditions, management, and amendment characteristics~\cite{zhang2025biochar}, while work on biochar interlayers examines how placement affects evaporation and surface salt accumulation~\cite{xu2023interlayer}. Studies of combined amendments also evaluate soil recovery together with crop performance and biological indicators~\cite{alcivar2018quinoa,rezapour2023combining}. Such evidence is more useful for planning when the intervention is represented by its rate, placement, water regime, and observation period rather than its name alone.

Scientific papers contain much of this evidence, but the information is dispersed across experimental descriptions, treatment tables, crop measurements, and post-treatment soil analyses. Manual synthesis becomes demanding when several soil indicators and management options must be considered simultaneously. Financial constraints add a further requirement: a technically plausible intervention may exceed the available budget or depend on water, materials, and maintenance that are unavailable locally. Economic assessments of salt-induced degradation and restoration underscore the importance of costs and the time horizon of reclamation benefits~\cite{qadir2014economics}. Agricultural decision-support research highlights the importance of making interacting constraints and uncertainty explicit and presenting information in a form that agricultural users can inspect~\cite{barons2024decision,htun2022visual}. For reclamation planning, the central information need is a traceable relationship among site conditions, measures, expenditure, and anticipated responses.

Large language models (LLMs) provide a flexible interface for organizing heterogeneous agricultural knowledge. Recent work has investigated crop-science instruction following, plant-protection question answering, agricultural knowledge graphs, and vision--language diagnosis~\cite{zhang2024empowering,xiong2025enhancing,zhao2024knowledge,awais2025agrogpt,wang2024agri}. A recent review places these developments within the broader use of foundation models for agricultural tasks~\cite{yin2025foundation}. However, performance on agricultural questions or image-based diagnosis does not establish the ability to assemble a consistent reclamation plan. This task requires the model to connect multiple site indicators with intervention mechanisms and to distinguish published treatment responses from expectations for a new site.

Retrieval-augmented generation (RAG) supplies external evidence at generation time~\cite{lewis2020rag}. In a closely related application, Chen et al.~\cite{chen2025coffee} developed a specialized RAG system for Yunnan Arabica coffee cultivation using hybrid retrieval and reranking. This work illustrates how a domain knowledge resource can support an agricultural information service. Saline-alkali land reclamation introduces a different planning requirement: the output must integrate measures and their implementation costs for a land scenario described by multiple indicators. It also raises a component-level question: when a general-purpose model and an adapted model receive the same retrieved evidence, does adaptation still improve the resulting plan?

This study develops Wenyao to examine that question. A bilingual corpus of 8,664 research papers is processed into SA-Bench, a resource of 4,860 paper-level planning instances organized using a 110-indicator schema. The adapted model links land-condition analysis, measures, and reported effects; case retrieval then supplies evidence for budget-conditioned planning. The objectives are to (i) assess the reliability of literature-based evidence extraction; (ii) quantify the contribution of domain adaptation and explanatory supervision to source-supported measure selection; and (iii) assess the complementary contributions of adaptation and retrieval to plan quality under matched evidence access. Extraction evaluation, source-disjoint multiple-choice testing, component comparisons, and blind expert assessment address these objectives.

\section{Materials and Methods}

\subsection{System Overview and Planning Task}
\label{sec:framework}

\begin{figure}[htbp]
\centering
\includegraphics[width=1\linewidth]{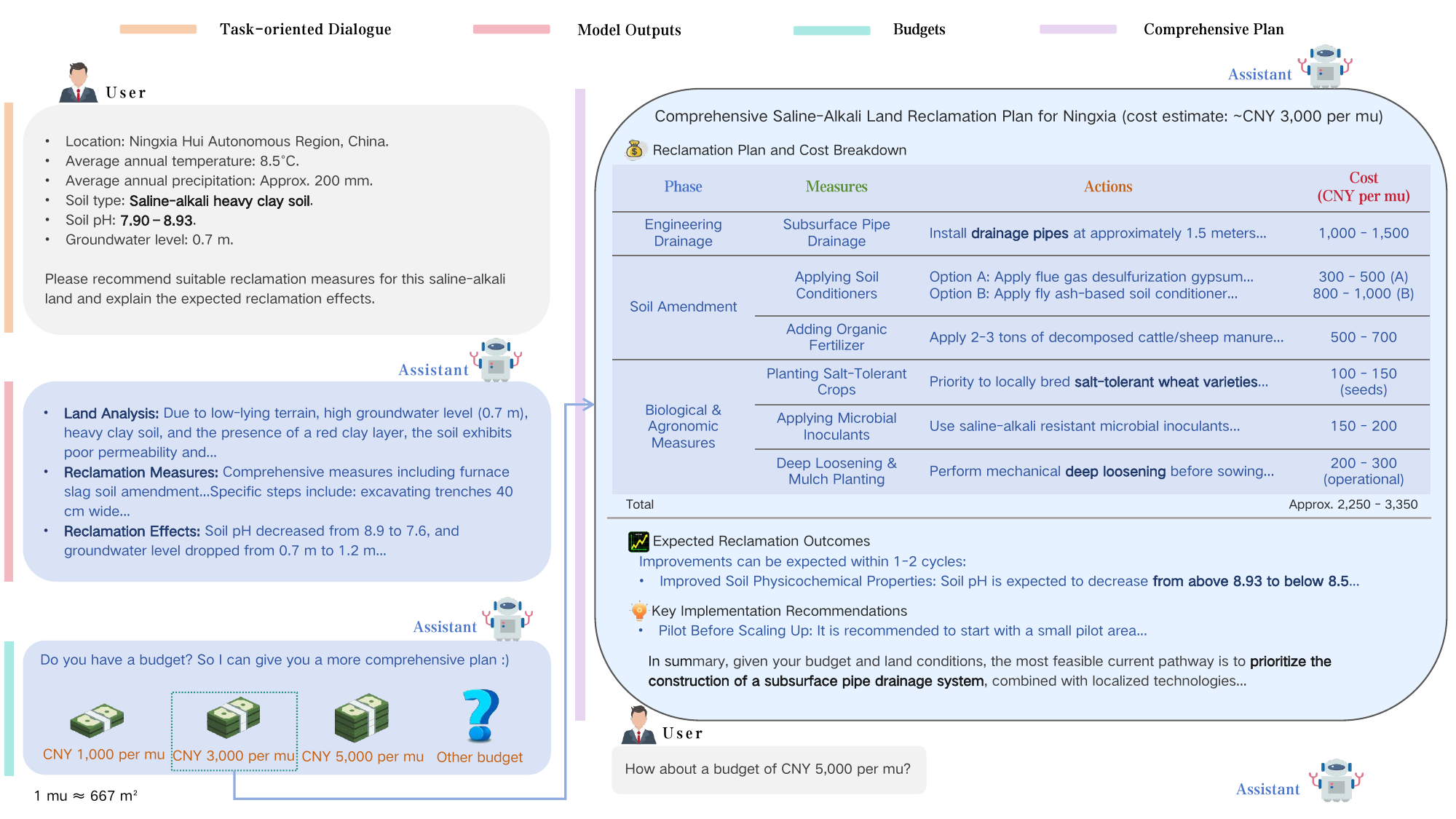}
\caption{System workflow connecting literature-derived evidence, model adaptation, site-specific analysis, case retrieval, and budget-aware planning. Multiple-choice evaluation uses the adapted model without the online retrieval branch. Only the training partition of SA-Bench is used for supervised adaptation.\label{fig:workflow}}
\end{figure}

Wenyao links scientific evidence with the soil, hydrological, climatic, and management information available for a target site. The offline workflow constructs SA-Bench and adapts a language model; the online workflow combines model-generated land analysis with retrieved cases and the user's budget (Figure~\ref{fig:workflow}). We use \emph{Wenyao} for the adapted model and \emph{Wenyao + RAG} for the complete planning system.

Given a site description $x$, the model generates
\begin{equation}
y=(a,m,e),
\end{equation}
where $a$ denotes land-condition analysis, $m$ the recommended measures, and $e$ the anticipated effects. The analysis connects observed site indicators with the proposed intervention mechanisms. For comprehensive planning, the user also supplies a budget $B$, and the system provides estimated expenditure and implementation considerations. Published treatment responses supply the evidence for anticipated effects, while expert review connects the generated recommendation with field implementation.

\subsection{Bilingual Literature and the Indicator Schema}

The initial corpus contains 8,664 research papers collected through Google Scholar, Web of Science, and CNKI: 4,486 English papers (51.8\%) and 4,178 Chinese papers (48.2\%). It covers reclamation measures, amendment strategies, crop selection, field experiments, and case studies. English-language sources include international study locations, while the separately collected planning cases concern 23 regions in China.

The extraction schema defines 110 indicator entries in six groups (Table~\ref{tab:schema}). In addition to initial soil conditions and interventions, the schema explicitly records crop growth and post-treatment responses, preserving the relationships needed for reclamation planning. The complete field catalogue is provided in Appendix~\ref{app:indicator_catalogue}, and the operative extraction instructions are summarized in Appendix~\ref{app:prompts}.

\begin{table}[htbp]
\caption{The six groups and 110 entries in the extraction schema. Counts refer to schema entries, including distinct pre- and post-treatment fields.\label{tab:schema}}
\small
\begin{tabularx}{\textwidth}{P{3.6cm}cL}
\toprule
\textbf{Group} & \textbf{Entries} & \textbf{Representative content}\\
\midrule
Experimental information & 5 & Location, coordinates, experiment type, meteorological conditions, and duration.\\
Initial land indicators & 35 & EC, salt content, pH and measurement method, ESP, SAR, ions, nutrients, texture, water status, and groundwater.\\
Measures and costs & 4 & Measure or amendment, application rate, timing, and cost.\\
Crop growth & 24 & Crop and cultivar, salt tolerance, physiology, canopy and root traits, water use, and nutrient or salt content.\\
Post-treatment effects & 37 & Biomass, yield, and post-treatment soil, water, ion, and nutrient indicators.\\
Comprehensive benefits & 5 & Yield increase, salinity or alkalinity reduction, input--output ratio, soil carbon, and microbial responses.\\
\midrule
\textbf{Total} & \textbf{110} & Conditions, treatments, crop responses, and outcomes.\\
\bottomrule
\end{tabularx}
\end{table}

\subsection{Extraction, Normalization, and Expert Quality Checks}

\begin{figure}[htbp]
\centering
\includegraphics[width=1\linewidth]{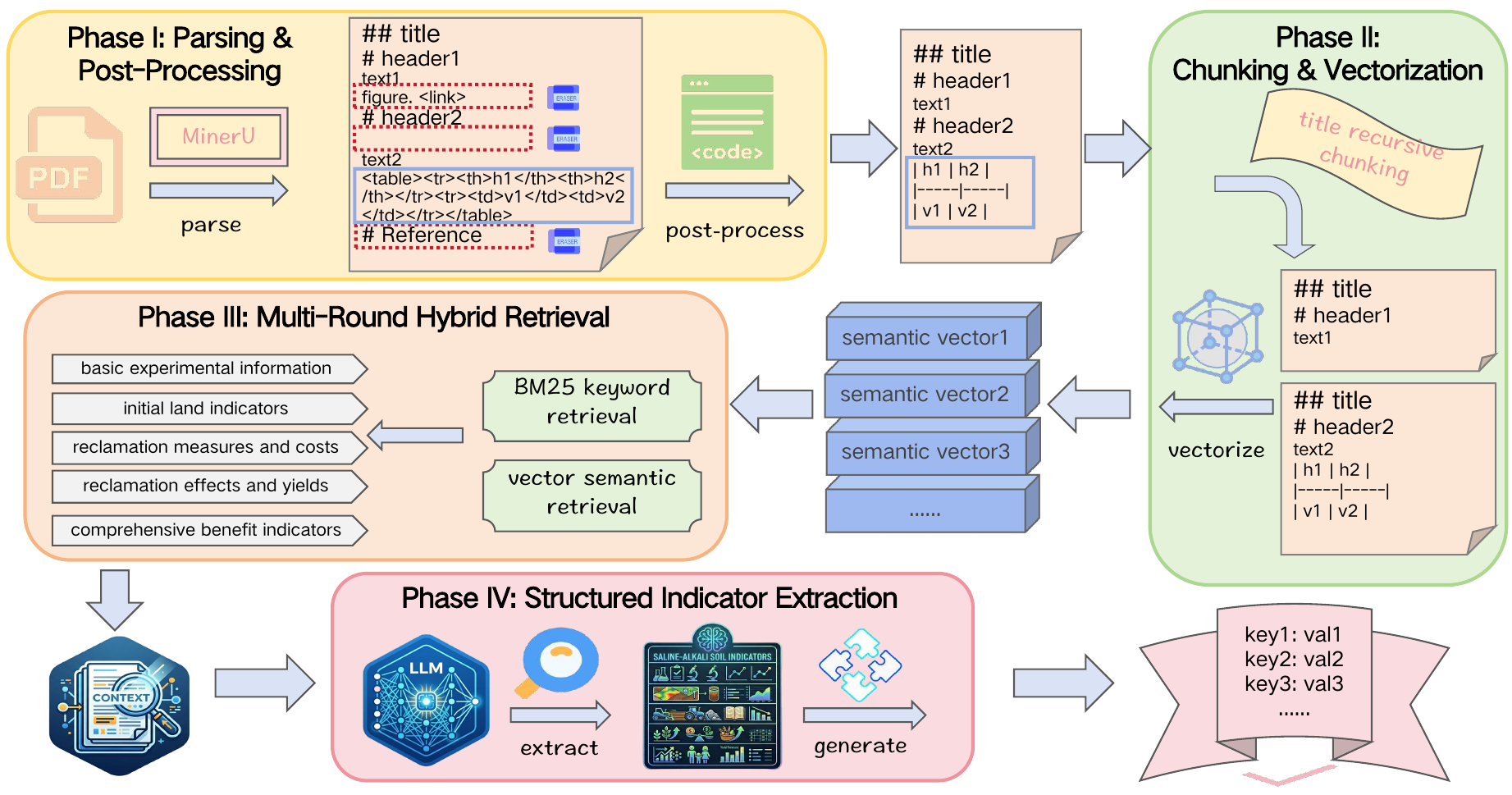}
\caption{Literature-to-evidence workflow. Documents shorter than 30,000 tokens supply full-text context; longer documents use heading-based hybrid retrieval and cross-encoder reranking. Six indicator groups guide structured extraction. Unit harmonization and expert checks support comparison with the original papers. The expert-checked sample contains 300 papers, and the retained resource contains 4,860 paper-level instances.\label{fig:pipeline}}
\end{figure}

The extraction workflow comprises document normalization, context-preserving indexing, grouped retrieval, constrained extraction, and expert checking (Figure~\ref{fig:pipeline}). MinerU~\cite{wang2024mineru} parses PDF text and tables. Post-processing removes redundant markup, normalizes tables into Markdown, and preserves the connection between table headers, treatment labels, and reported values.

Documents shorter than 30,000 tokenizer tokens supply their complete cleaned text as extraction context. Longer documents are divided at Markdown heading levels 1--3, retaining the headings to preserve section context. The workflow uses heading-defined sections rather than fixed-length overlapping chunks. It is implemented in LangChain 0.3.27; the component versions and configuration are reported in Appendix~\ref{app:implementation}.

BGE-M3~\cite{chen2024bge} produces normalized dense embeddings for a Chroma index. BM25 supports lexical matching~\cite{robertson2009probabilistic}. Dense retrieval returns ten candidates, and The lexical and dense rankings are combined by weighted reciprocal rank fusion (RRF)~\cite{cormack2009rrf}:
\begin{equation}
R(d,q)=\sum_{j\in\{\mathrm{lex},\mathrm{dense}\}}
\frac{w_j\,\mathbf{1}[d\in\mathcal{D}_j(q)]}{60+r_j(d,q)},
\qquad w_{\mathrm{lex}}=w_{\mathrm{dense}}=0.5,
\end{equation}
where $r_j$ is the one-based rank in channel $j$. The constant 60 is the default in LangChain 0.3.27, and repeated document contents are merged. Lexical and semantic retrieval provide complementary evidence coverage~\cite{kuzi2020leveraging}.

For long documents, each indicator group is queried separately because its evidence may occur in different sections or tables. A cross-encoder reranker, \nolinkurl{BAAI/bge-reranker-v2-m3}, reranks the fused candidates and retains up to ten chunks per group:
\begin{equation}
\mathcal{C}_g=\operatorname{Top}^{\mathrm{cross}}_{10}
\left(\operatorname{RRF}_{0.5,0.5}
\left[\operatorname{BM25}_{k_B}(q_g),\operatorname{Dense}_{10}(q_g)\right]\right).
\end{equation}
The implementation uses the BM25 instance default $k_B=4$. Software versions and the distinction between this extraction path and online case retrieval are documented in Appendix~\ref{app:implementation}.

The extraction experiments use DeepSeek-R1-Distill-Qwen-32B to produce JSON records from the group-specific fields and selected context. The prompt retains multiple treatment rows and associated explanations, fixes the order of output fields, and represents unreported values as \texttt{Not provided}. The six field groups and the extraction constraints are documented in Appendix~\ref{app:prompts}.

Extracted indicators are harmonized to common units and normalized representations before comparison with expert-checked reference values. Measurement methods and treatment context are retained with the values. Domain experts inspect a random sample of 300 papers. Numerical indicators are compared directly after unit and representation normalization, while textual indicators are judged for semantic equivalence by the experts. For each paper, precision is the fraction of extracted indicators judged correct, recall is the fraction of reference indicators recovered correctly, and F1 is their harmonic mean. These metrics are averaged across papers. Paper-level bootstrap resampling with replacement uses 1,000 replicates~\cite{efron1979bootstrap}; each replicate resamples the 300 annotated papers and recomputes the paper-averaged metrics. The resulting empirical distribution provides the reported 95\% bootstrap intervals.

\subsection{Paper-Level Instances and Source-Disjoint Partitioning}

After curation, 4,860 source papers each contribute one planning instance and one corresponding MCQ. Each instance combines the initial land description with an explanatory analysis, supported measures, and reported effects. Cleaning and deduplication retain records containing more than 20 indicators. MCQs pair the source-supported measure with three alternatives drawn from other instances. Their answers are defined by the intervention supported in the corresponding source case. Document-derived questions provide a task-specific assessment of the recovered knowledge~\cite{guinet2024exam}; they do not establish that the keyed measure is the only agronomically feasible intervention.

Papers are randomly assigned to disjoint training and test partitions. No information derived from a test paper appears in the supervised training partition. The training and test partitions contain 3,924 and 936 instances, corresponding to 80.74\% and 19.26\% of the 4,860-paper resource, respectively.

\begin{table}[htbp]
\caption{Paper-level corpus, training/test counts, and instance-length statistics. Each retained source paper contributes one instance.\label{tab:dataset}}
\begin{tabularx}{\textwidth}{Lr}
\toprule
\textbf{Statistic} & \textbf{Value}\\
\midrule
Initial source papers & 8,664\\
English / Chinese source papers & 4,486 / 4,178\\
Retained papers / planning instances & 4,860 / 4,860\\
Training instances / held-out MCQs & 3,924 / 936\\
Input tokens, minimum / mean / maximum & 66 / 453 / 1,605\\
Output tokens, minimum / mean / maximum & 291 / 1,194 / 2,862\\
Total tokens, minimum / mean / maximum & 385 / 1,647 / 4,020\\
\bottomrule
\end{tabularx}
\end{table}

\begin{figure}[tbp]
\centering
\includegraphics[width=\textwidth]{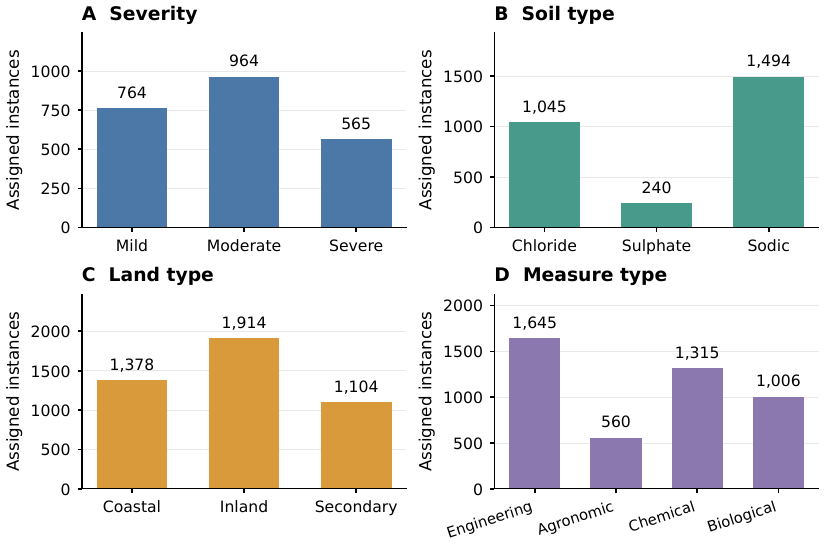}
\caption{Assigned SA-Bench labels across salinity severity, soil type, land type, and intervention category. Undetermined labels are not plotted, so the totals vary across dimensions. These are full-resource counts; held-out category counts appear in Appendix~\ref{app:categories}.\label{fig:distribution}}
\end{figure}

The retained cases cover engineering, agronomic, chemical, and biological measures. Appendix~\ref{app:example} illustrates how site conditions, interventions, and reported responses are organized.

\subsection{Domain Adaptation and Greedy Generation}

Wenyao is initialized from QwQ-32B~\cite{qwen2025qwq} and adapted through LoRA~\cite{hu2022lora} in LLaMA-Factory~\cite{zheng2024llamafactory}. The training mixture adds 2,000 general examples from \texttt{alpaca-gpt4-}\allowbreak\texttt{data-zh} and 94 identity examples to the source-disjoint SA-Bench training partition. Targets contain land analysis, measures, and effects. The supervised objective is
\begin{equation}
\mathcal{L}_{\mathrm{SFT}}=-\sum_i\sum_t
\log p_{\theta,\phi}(y_{i,t}\mid x_i,y_{i,<t}),
\end{equation}
where $\theta$ denotes the base model and $\phi$ the trainable LoRA parameters. The rationale ablation removes the land analysis and explanatory content while retaining the same scenarios, measures, and effects.

Training uses two H20 GPUs with 96 GB each, 32 CPU processors, and 192 GB system memory. LoRA uses rank 8, alpha 16, dropout 0, and the attention projection modules \nolinkurl{q_proj}, \nolinkurl{k_proj}, \nolinkurl{v_proj}, and \nolinkurl{o_proj}. The remaining adaptation settings appear in Appendix~\ref{app:implementation}; unspecified options use the framework defaults. Inference uses greedy generation: sampling is disabled, and API-based generation uses temperature zero. The MCQ evaluations use the same scenarios and answer options for all models without retrieved evidence.

\subsection{Budget-Aware Retrieval-Augmented Planning}

Wenyao first analyzes the target land and proposes candidate measures. Online retrieval concatenates the land request and initial response as its query. Each evidence-store document contains a case instruction and response, with its source-paper identity retained as metadata. The planner uses the same normalized BGE-M3 embeddings, Chroma dense retrieval with $k=10$, BM25, and equal-weight RRF with constant 60. The online planning workflow directly uses the fused case list without the additional cross-encoder stage used for long-document extraction. Cost information is assembled for the retrieved measures before budget-conditioned generation. The final generation stage integrates the land information, initial analysis, retrieved evidence, and budget:
\begin{equation}
\hat y_B=G_{\theta,\phi}
\left(x,B,\hat y_0,\mathcal{E}(x,\hat y_0)\right).
\end{equation}
Here, $\hat y_0$ is the initial model output and $\mathcal{E}$ denotes retrieved evidence. The resulting plan organizes proposed operations, estimated expenditure, and anticipated effects for expert review.

For matched-evidence evaluation, Qwen3-Max receives the same retrieved cases as Wenyao. This separates the contribution of evidence access from the additional contribution of the adapted generator. The fused list is deduplicated by content and has no additional fixed top-$n$ cutoff in the planning implementation; its size depends on overlap between the two channels.

\subsection{Evaluation Protocols}

\subsubsection{Model-Level Measure Selection}

The comparison models are QwQ-32B, Qwen3-Max, DeepSeek-V3, DeepSeek-R1, GPT-5, and Gemini-2.5-Pro~\cite{liu2024deepseek,guo2025deepseek,comanici2025gemini}. Each receives the same held-out scenario and four candidate measures under greedy decoding. Accuracy is the proportion of the 936 questions for which the selected option matches the source-supported answer. Mean wall-clock response time is recorded over the same task.

\subsubsection{Automatic Plan Comparison over 69 Cases}

The practical scenario pool covers 23 regions under three budgets: 1,000, 3,000, and 5,000 \CNYmu. One mu is approximately 667 m$^2$. This yields 69 region--budget cases for each system. DeepSeek-R1 and GPT-5 compare paired plans on precision, expertise, practicality, and innovation and select a preferred plan. Self-comparisons are omitted. Model-based judging provides a complementary assessment of generated outputs~\cite{zheng2023judge}; the expert assessments described below provide the domain-specific human evaluation.

For a system pair, the first system's preference rate is $100W/69$, where $W$ is the number of cases in which the judge prefers it. Preference rates are displayed as percentages rounded to two decimal places. The percentages are calculated from the integer preference counts using the common denominator of 69. These are relative plan-preference rates rather than measurements of field effectiveness.

\subsubsection{Expert Assessment and Score Aggregation}

The primary human assessment selects 12 of the same 23 regions and includes all three budgets for each, giving 36 questionnaires drawn from the 69-case automatic pool. Each questionnaire contains four anonymized plans, in randomized order, from Wenyao + RAG, Qwen3-Max, DeepSeek-R1, and GPT-5. Thirty-one experts from 22 institutes participate, with a mean reported domain experience of 7.7 years. Five or six experts assess each questionnaire.

Experts assign 0--10 scores for rationality of objectives and measures (\emph{Objectives}), accuracy of cost estimates (\emph{Cost}), and validity of projected outcomes (\emph{Conclusions}). Multiple ratings of the same questionnaire are averaged before system comparison:
\begin{equation}
\bar s_{qmc}=\frac{1}{K_q}\sum_{r=1}^{K_q}s_{qrmc},
\end{equation}
where $q$ indexes a region--budget questionnaire, $r$ an expert, $m$ a model, and $c$ a criterion. Individual expert ratings are retained separately for inter-rater reliability estimation.

A supplementary matched-evidence assessment compares Wenyao, Wenyao + RAG, Qwen3-Max, and Qwen3-Max + RAG using six experts, eight regions, and the same three budgets, giving 24 cases. It is reported separately from the primary assessment.

\subsubsection{Statistical Analysis}

Inter-rater reliability is evaluated from the individual expert ratings using a two-way random-effects, absolute-agreement ICC for average ratings, ICC(2,$k$), where $k$ is the number of ratings in the corresponding expert panel~\cite{koo2016icc}. For model comparisons, expert ratings are first averaged within each questionnaire, and the three budget-specific means are then averaged within a region. The region is the independent paired unit, giving 12 pairs in the primary assessment and eight in the supplementary assessment. Two-sided paired Wilcoxon signed-rank tests compare Wenyao + RAG with each baseline on each criterion. Holm adjustment~\cite{holm1979multiple} is applied to the three primary baseline contrasts within a criterion, with an adjusted significance threshold of 0.05.

Paired Cohen's $d_z$ is the mean of the regional score differences divided by their standard deviation. Magnitudes of 0.2, 0.5, and 0.8 are used as conventional small, medium, and large interpretive references~\cite{cohen1988power}; the standardization uses paired differences.

\section{Results}

\subsection{Evidence-Extraction Performance}

On 300 expert-annotated papers, the full pipeline achieves precision 0.76, recall 0.92, and F1 0.83, with a 95\% bootstrap interval of [0.80, 0.85] for F1 (Table~\ref{tab:extraction}). Evaluation compares normalized indicators after unit harmonization, using 1,000 bootstrap replicates.

\begin{table}[htbp]
\caption{Evidence-extraction ablations on 300 expert-annotated papers. Intervals are the reported 95\% bootstrap intervals for F1.\label{tab:extraction}}
\small
\setlength{\tabcolsep}{4pt}
\begin{tabularx}{\textwidth}{Lcccc}
\toprule
\textbf{Configuration} & \textbf{Precision} & \textbf{Recall} & \textbf{F1} & \textbf{95\% CI}\\
\midrule
Full pipeline & \textbf{0.76} & \textbf{0.92} & \textbf{0.83} & [0.80, 0.85]\\
pdfplumber only & 0.60 & 0.62 & 0.61 & [0.58, 0.63]\\
BM25 retrieval only & 0.72 & 0.85 & 0.78 & [0.75, 0.80]\\
Dense retrieval only & 0.70 & 0.83 & 0.77 & [0.74, 0.79]\\
Without multi-round batching & 0.74 & 0.76 & 0.75 & [0.73, 0.78]\\
Simplified extraction prompts & 0.66 & 0.90 & 0.76 & [0.75, 0.78]\\
\bottomrule
\end{tabularx}
\end{table}

Replacing the parsing and normalization stage with pdfplumber alone reduces F1 from 0.83 to 0.61. Hybrid retrieval outperforms either retrieval channel alone, while grouped retrieval improves recall by collecting evidence for indicators distributed across sections and tables. Structured prompts improve precision relative to simplified instructions. Together, these ablations support preserving document structure and aligning extraction with the indicator schema.

\subsection{Measure Selection on Source-Disjoint Papers}

\begin{figure}[htbp]
\centering
\includegraphics[width=\textwidth]{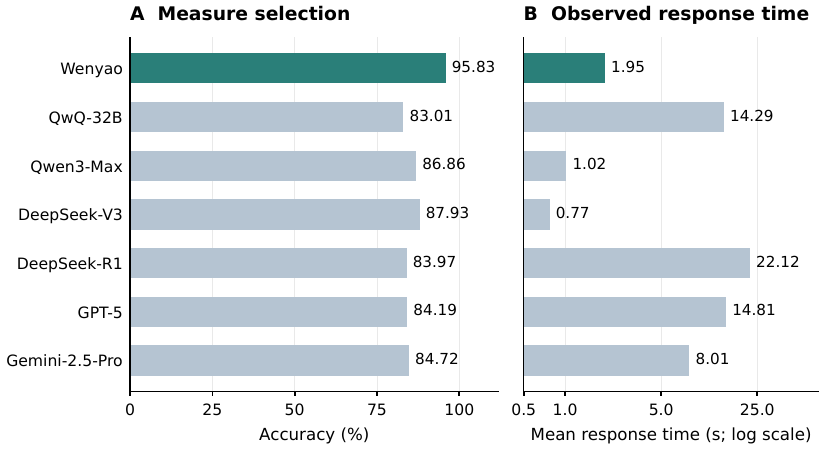}
\caption{Multiple-choice accuracy and mean response time on 936 held-out questions without retrieval. Response times describe the reported serving environments; the time axis is logarithmic.\label{fig:model_results}}
\end{figure}

Wenyao achieves 95.83\% accuracy on the 936 held-out MCQs, exceeding DeepSeek-V3, the strongest general-purpose baseline, by 7.90 percentage points (87.93\%). QwQ-32B attains 83.01\%, giving a 12.82-point improvement after adaptation (Figure~\ref{fig:model_results}). All models receive identical scenarios and candidate options without retrieval.

Category-level results appear in Appendix~\ref{app:categories}. Against DeepSeek-V3, Wenyao's improvements are 1.50 points for engineering measures, 11.93 for agronomic measures, 10.73 for chemical measures, and 12.87 for biological measures. The larger gains in agronomic, chemical, and biological categories indicate improved recognition of the source-supported measures represented in SA-Bench. Wenyao's observed mean response time is 1.95 s per question, compared with 14.29 s for QwQ-32B. Because these observations use the reported serving environments, they do not isolate a hardware-controlled efficiency advantage.

\subsection{Rationale-Enhanced Supervision}

Removing the land analysis and explanatory content from the training targets reduces overall accuracy from 95.83\% to 89.42\%, a decrease of 6.41 percentage points. Inputs, source cases, target measures, and effects remain fixed. The result supports the contribution of explanatory supervision to learning the relationships among site conditions, measures, and outcomes.

\subsection{Retrieval and Adaptation in Comprehensive Planning}

Automatic comparisons over all 69 region--budget cases favor Wenyao + RAG over the baseline models in the valid comparisons (Appendix~\ref{app:original_comparisons}). Table~\ref{tab:matched_rag} separates the effects of adaptation and evidence access by comparing Wenyao and Qwen3-Max with and without the same retrieved cases.

\begin{table}[htbp]
\caption{Preference rates (\%) favoring the first system. Each judge evaluates the same 69 region--budget cases; values are $100W/69$, rounded to two decimal places.\label{tab:matched_rag}}
\small
\begin{tabularx}{\textwidth}{Lcc}
\toprule
\textbf{First system versus second system} & \textbf{DeepSeek-R1} & \textbf{GPT-5}\\
\midrule
Wenyao versus Qwen3-Max & 65.22 & 62.32\\
Wenyao versus Qwen3-Max + RAG & 55.07 & 52.17\\
Wenyao + RAG versus Wenyao & 66.67 & 60.87\\
Wenyao + RAG versus Qwen3-Max & 76.81 & 75.36\\
Wenyao + RAG versus Qwen3-Max + RAG & 60.87 & 57.97\\
Qwen3-Max + RAG versus Qwen3-Max & 72.46 & 69.57\\
\bottomrule
\end{tabularx}
\end{table}

Adding retrieval to Wenyao receives preference rates of 66.67\% from DeepSeek-R1 and 60.87\% from GPT-5. Under matched evidence access, the adapted generator also retains a majority preference. The comparison with Qwen3-Max follows the same complementary pattern: retrieval strengthens the general-purpose generator, while the domain-adapted model provides further benefits in organizing a condition-specific plan.

\subsection{Blind Expert Assessment}

In the primary study, Wenyao + RAG receives the highest mean scores for Objectives (6.89) and Conclusions (7.60; Table~\ref{tab:experts_primary}). Scores are first averaged across the experts assigned to a questionnaire and then summarized across the evaluated cases.

\begin{table}[htbp]
\caption{Primary assessment by 31 experts from 22 institutes. The 36 questionnaires cover 12 regions sampled from the 23-region automatic evaluation pool, with three budgets per region. Each questionnaire has five or six expert ratings on a 0--10 scale.\label{tab:experts_primary}}
\begin{tabularx}{\textwidth}{Lccc}
\toprule
\textbf{System} & \textbf{Objectives} & \textbf{Cost} & \textbf{Conclusions}\\
\midrule
Wenyao + RAG & \textbf{6.89} & 6.95 & \textbf{7.60}\\
Qwen3-Max & 6.80 & 6.98 & 6.80\\
DeepSeek-R1 & 6.55 & 7.00 & 7.10\\
GPT-5 & 6.79 & \textbf{7.13} & 7.43\\
\bottomrule
\end{tabularx}
\end{table}

Inter-rater ICC(2,$k$) estimates range from 0.68 to 0.95. The regional paired analysis identifies statistically significant system differences under the Wilcoxon protocol with Holm adjustment ($p_{\mathrm{adj}}<0.05$), with paired effect magnitudes of approximately $|d_z|=0.8$. These aggregate inferential summaries accompany the criterion-specific mean scores in Table~\ref{tab:experts_primary} and should not be interpreted as evidence that every system--criterion contrast is significant. GPT-5 obtains the highest mean Cost score (7.13), whereas Wenyao + RAG scores 6.95 on this criterion. 

The supplementary six-expert study evaluates both generators under matched evidence access (Table~\ref{tab:experts_matched}). Retrieval increases all three mean scores for each generator. Wenyao + RAG attains 7.16 for Objectives and 7.68 for Conclusions, compared with 6.96 and 7.38 for Qwen3-Max + RAG. The descriptive differences support a contribution from adaptation to objectives and projected outcomes under common evidence access. Qwen3-Max + RAG receives the highest mean Cost score (7.02), compared with 6.92 for Wenyao + RAG.

\begin{table}[htbp]
\caption{Supplementary matched-evidence assessment by six experts over eight regions and three budgets, giving 24 region--budget cases. Means are reported separately from the primary study.\label{tab:experts_matched}}
\begin{tabularx}{\textwidth}{Lccc}
\toprule
\textbf{System} & \textbf{Objectives} & \textbf{Cost} & \textbf{Conclusions}\\
\midrule
Wenyao & 6.82 & 6.63 & 6.90\\
Wenyao + RAG & \textbf{7.16} & 6.92 & \textbf{7.68}\\
Qwen3-Max & 6.67 & 6.94 & 6.86\\
Qwen3-Max + RAG & 6.96 & \textbf{7.02} & 7.38\\
\bottomrule
\end{tabularx}
\end{table}

\section{Discussion}

\subsection{Agronomic Interpretation of Evidence-Based Planning}

The results indicate that domain adaptation and case retrieval can help organize the information needed for reclamation planning. The agricultural value lies in connecting site conditions with the mechanisms and implementation requirements of a proposed measure. The 110-indicator schema records initial conditions, treatment details, crop growth, and post-treatment responses in a common structure. This organization is consistent with the integrated view of salt-affected soil management presented in reclamation reviews~\cite{shaygan2022reclamation,wang2024saline,zhu2024health}.

Experimental studies of leaching with organic amendments and of gypsum--compost--water combinations show why water quality and treatment interactions must accompany the amendment description~\cite{chaganti2015leaching,sundha2020conjunctive}. This provides an agronomic rationale for retaining treatment and hydrological context in the evidence records.

The intervention-category results suggest that adaptation was particularly helpful for recognizing source-supported agronomic, chemical, and biological measures. These choices often depend on combinations of indicators rather than a single threshold. The rationale ablation further suggests that explanatory supervision helped the model learn relationships among conditions, measures, and reported effects. The gains concern source-supported measure selection and plan presentation.

\subsection{Complementary Roles of Adaptation and Retrieval}

The matched-evidence comparisons help distinguish the contribution of an adapted generator from the contribution of access to external cases. Retrieval improved plan preference and expert mean scores for both Wenyao and Qwen3-Max. Wenyao with retrieval retained higher mean scores for objectives and projected outcomes when the comparator received the same cases. These findings support the combination of domain-specific training and case-based evidence access for this planning task, without implying that adaptation replaces verification of the retrieved material.

The coffee-cultivation system of Chen et al.~\cite{chen2025coffee} provides a relevant comparison: both applications use a specialized evidence resource to support agricultural knowledge services. The present study examines planning across soil indicators and budget levels and compares adapted and general-purpose generators under shared evidence access. Its extraction F1, measure-selection accuracy, and plan-preference rates address different evaluation targets from the coffee study's citation-level F1 and therefore should not be compared numerically. More generally, research on long-context language models shows that including relevant information does not guarantee reliable use of that information~\cite{liu2024lost}. The extraction ablations here support retaining document structure and selecting context by indicator group.

\subsection{Practical Implications for Agricultural Advisers}

Wenyao is most appropriately used to prepare a plan for professional review. The output organizes land analysis, proposed operations, estimated expenditure, and anticipated responses, enabling an adviser to inspect the links between the request and the available evidence. This role aligns with agricultural decision-support research that emphasizes user interpretation and the treatment of interacting constraints~\cite{htun2022visual,barons2024decision}. Relevant source conditions, measurement methods, treatment rates, and observation periods remain necessary when deciding whether a retrieved case is applicable.

Budget conditioning should also be distinguished from validated economic optimization. In the primary expert study, GPT-5 obtained the highest mean Cost score; in the supplementary matched-evidence study, Qwen3-Max with retrieval obtained the highest score on that criterion. The present system consequently has stronger support for organizing objectives and projected responses than for improving cost accuracy. Local material prices, labor, transport, irrigation availability, and maintenance requirements must be checked before implementation. The three evaluated budgets represent planning constraints, and the reported scores do not measure realized expenditure, profitability, or long-term maintenance costs.

\subsection{Limitations of the Evidence and Evaluation}

Several limitations affect the interpretation of the results. First, extraction precision of 0.76 indicates that errors remain despite high recall. Unit harmonization cannot remove differences in sampling depth, EC or pH measurement method, experimental duration, or treatment context. Retrieved values and the relationships among treatment rows require verification against the original paper. The evidence resource reflects the available published literature and its reporting practices rather than a prospectively balanced set of field trials.

Second, paper-level separation prevents supervised training on the held-out source papers but does not establish separation by site, region, study family, or foundation-model pretraining data. The MCQs assess recognition of the recorded intervention among alternatives drawn from other cases. A source-supported answer need not be the only agronomically feasible option. The 23-region planning pool and both expert assessments concern China; a bilingual source corpus alone does not demonstrate transfer to other agricultural regions.

Third, automated plan preferences and expert ratings assess proposed plans rather than observed crop or soil responses. Model judges can exhibit position, verbosity, and self-enhancement biases~\cite{zheng2023judge}. The human assessments provide complementary professional judgment, but their independent regional sample sizes are 12 and eight, respectively. The aggregate statistical summaries do not resolve every criterion-by-baseline contrast, and the small difference in the primary Objectives means should be interpreted alongside that limitation. Likewise, response times measured in different serving environments do not establish a controlled computational-efficiency advantage.

Finally, retrieval access does not ensure that every generated claim is supported by a source. Work on citation-based generation and RAG evaluation distinguishes evidence retrieval, answer relevance, and claim support~\cite{gao2023citations,es2024ragas}. Those distinctions motivate future evaluation of source attribution and unsupported recommendations in addition to overall plan preference. The present study does not report a claim-level citation audit or a prospective agronomic outcome comparison.

\subsection{Field Validation and Further Development}

Following technical review by domain experts, field validation has been initiated, with soil and crop observations ongoing over the relevant growth cycles. No completed field-treatment outcome data are reported here. The next stage should relate the implemented measures to changes in soil conditions, crop establishment and yield, and actual expenditure, while recording the local conditions that affect transfer from a published case. The planned release of derived data, implementation code, indicator specifications, and evaluation questionnaires will support replication. Further development should prioritize evidence quality, current local cost information, and outcome-based validation before broader deployment.

\section{Conclusions}

Integrating domain adaptation with case retrieval provided a literature-informed approach to saline-alkali land reclamation planning. The 110-indicator extraction workflow achieved F1 0.83 on 300 expert-annotated papers, and Wenyao achieved 95.83\% source-supported measure-selection accuracy on 936 held-out questions without retrieval. Removing explanatory supervision reduced accuracy to 89.42\%. Matched-evidence comparisons supported complementary contributions from adaptation and retrieval. In the expert assessments, the complete system obtained the highest mean scores for objectives and projected outcomes, while the strongest cost scores were obtained by comparator systems. These results support the use of Wenyao to organize reclamation evidence and prepare plans for professional review. Prospective validation of soil improvement, crop response, and realized costs remains necessary before judging agronomic or economic effectiveness.

\Needspace{10\baselineskip}
\section*{Declarations}
\declaration{Author contributions.}{\ContributionStatement}
\declaration{Funding.}{\FundingStatement}
\declaration{Ethics statement.}{\EthicsStatement}
\declaration{Informed consent.}{\ConsentStatement}
\declaration{Data availability.}{\DataAvailabilityStatement}
\declaration{Acknowledgments.}{We thank the domain experts who contributed to evidence checking and blind assessment of reclamation plans. Generative models were used for evidence extraction, instruction-data construction, plan generation, and complementary automatic evaluation as described in the Methods.}
\declaration{Conflicts of interest.}{\ConflictStatement}

\section*{Abbreviations}
\noindent\begin{tabularx}{\textwidth}{@{}lX@{}}
BM25 & Best Matching 25\\
EC & Electrical conductivity\\
ESP & Exchangeable sodium percentage\\
ICC & Intraclass correlation coefficient\\
LLM & Large language model\\
LoRA & Low-rank adaptation\\
MCQ & Multiple-choice question\\
RAG & Retrieval-augmented generation\\
SAR & Sodium adsorption ratio\\
SFT & Supervised fine-tuning
\end{tabularx}

\appendix
\setcounter{table}{0}
\renewcommand{\thetable}{A\arabic{table}}
\section{Implementation and Prompts}
\label{app:implementation}

Table~\ref{tab:configuration} records the adaptation, greedy decoding, and extraction-evaluation settings. Table~\ref{tab:retrieval_configuration} provides the retrieval and reranking implementation settings, with software versions recorded in the implementation specification.

\begin{table}[htbp]
\caption{Adaptation, generation, and extraction-evaluation settings.\label{tab:configuration}}
\small
\begin{tabularx}{\textwidth}{P{4.0cm}L}
\toprule
\textbf{Component} & \textbf{Reported setting}\\
\midrule
Structured extraction model & DeepSeek-R1-Distill-Qwen-32B\\
Planning model initialization & QwQ-32B\\
Framework / adaptation & LLaMA-Factory; LoRA\\
Learning rate / optimizer & $5\times10^{-5}$ / AdamW\\
Epochs / LoRA rank & 6 / 8\\
LoRA alpha / dropout & 16 / 0\\
LoRA target modules & \nolinkurl{q_proj}, \nolinkurl{k_proj}, \nolinkurl{v_proj}, \nolinkurl{o_proj}\\
Per-device batch / accumulation & 4 / 8\\
Auxiliary general / identity examples & 2,000 / 94\\
Generation & Greedy; \texttt{do\_sample=false}; temperature 0\\
Extraction bootstrap replicates & 1,000\\
Hardware & 2 H20 GPUs, 96 GB each; 32 CPUs; 192 GB RAM\\
\bottomrule
\end{tabularx}
\end{table}

\begin{table}[htbp]
\caption{Retrieval and reranking settings. Library defaults correspond to the reported software versions. The extraction and planning paths have different final retrieval stages.\label{tab:retrieval_configuration}}
\small
\begin{tabularx}{\textwidth}{P{4.0cm}L}
\toprule
\textbf{Component} & \textbf{Implementation setting}\\
\midrule
LangChain / community & 0.3.27 / 0.3.29\\
Chroma / Hugging Face integrations & \nolinkurl{langchain_chroma} 0.2.6; \nolinkurl{langchain_huggingface} 0.3.1\\
Core / text splitters & 0.3.76 / 0.3.11\\
Sentence Transformers / rank-bm25 & 5.0.0 / 0.2.2\\
Embedding model & \nolinkurl{BAAI/bge-m3}; CUDA; normalized embeddings\\
Full-text extraction threshold & Fewer than 30,000 tokenizer tokens\\
Long-document splitting & Markdown headings 1--3; \texttt{strip\_headers=false}\\
Dense retrieval & Chroma; \texttt{search\_kwargs=\{"k":10\}}\\
BM25 lexical retrieval & Instance default $k_B=4$; default whitespace tokenization\\
Fusion & Weighted RRF; weights $(0.5,0.5)$; default constant $c=60$; content-based deduplication\\
Extraction reranker & \nolinkurl{BAAI/bge-reranker-v2-m3}; \texttt{CrossEncoderReranker}; explicit \texttt{top\_n=10}\\
Other reranker model options & Pretrained-model and wrapper defaults\\
Online planning retrieval & Case-level RRF output; no subsequent cross-encoder or fixed final cutoff\\
Indicator extraction & Six groups, 110 entries; maximum five JSON/field-validation attempts\\
\bottomrule
\end{tabularx}
\end{table}

\subsection{Structured Indicator Extraction Prompt}
\label{app:prompts}

The following excerpt summarizes the operative constraints of the extraction specification. The field list and evidence context are populated separately for each of the six indicator groups. The complete field catalogue is reproduced below.

\begin{quote}
You are an agricultural expert. Please strictly extract relevant indicators for saline-alkali land remediation based on the following context.\par
Requirements: Extract information strictly from the context. Fabrication, inference, or supplementation is prohibited. For tabular data, completely extract all relevant values; retain all relevant rows. For indicators not explicitly mentioned, return ``Not provided''. Retain explanatory context needed to understand an indicator. The number and order of output fields must match the given indicator list. Return JSON directly, without additional explanations or a code-fence prefix.\par
Indicator list to extract: \texttt{\{metrics\}}\par
Context: \texttt{\{context\}}
\end{quote}

An illustrative output structure is shown below. The placeholders describe schema fields and are not additional experimental observations.
\par\noindent\begin{minipage}{\linewidth}
\begin{lstlisting}
{
  "Location": "value explicitly stated in context",
  "Soil pH": "value explicitly stated in context",
  "Reclamation measures": "reported treatments",
  "Reclamation effects": "reported outcomes",
  "Unreported indicator": "Not provided"
}
\end{lstlisting}
\end{minipage}\par

\subsection{Complete Indicator Catalogue}
\label{app:indicator_catalogue}
The extraction schema contains 110 entries. The catalogue below reproduces their field names; the pre- and post-treatment prefixes are omitted within the corresponding groups for readability. The extraction constraints are summarized in Appendix~\ref{app:prompts}. Field names follow the extraction schema and should be interpreted with the measurement methods and treatment context reported in each source paper. Prompt examples illustrate the extraction format and are not additional experimental observations.
\begingroup\small
\noindent\textbf{Experimental information (5 entries).} Location; Latitude and Longitude; Experiment Type; Meteorological Conditions; Experiment Duration.\par\medskip

\noindent\textbf{Initial land indicators (35 entries).} Saline-Alkali Land Type; Electrical Conductivity (EC); Total Soluble Salt Content; pH; pH Measurement Method; Exchangeable Sodium Percentage (ESP); Sodium Adsorption Ratio (SAR):; Cation Exchange Capacity (CEC); Sodium Ion (Na$^{+}$) Content; Calcium Ion (Ca$^{2+}$) Content; Magnesium Ion (Mg$^{2+}$) Content; Chloride Ion (Cl$^{-}$) Content; Sulfate Ion (SO$_{4}$$^{2-}$) Content; Bicarbonate (HCO$_{3}$$^{-}$) Content; Carbonate (CO$_{3}$$^{2-}$) Content; Soil Organic Matter (SOM) Content; Soil Texture; Soil Bulk Density; Total Porosity; Soil Temperature; Calcium Carbonate (CaCO$_{3}$) Content; Available Phosphorus (P) Content; Available Potassium (K) Content; Total Nitrogen (N) Content; Ammonium Nitrogen (NH$_{4}$$^{+}$-N) Content; Nitrate Nitrogen (NO$_{3}$$^{-}$-N) Content; Trace Element Content; Field Capacity (FC); Permanent Wilting Point (PWP); Soil Water Content (SWC); Saturated Hydraulic Conductivity (Ks); Capillary Rise Height; Groundwater Depth; Irrigation Water EC; Annual Evaporation/Precipitation Ratio.\par\medskip

\noindent\textbf{Measures and costs (4 entries).} Measure/Amendment; Application Rate/Intensity; Duration/Timing; Cost.\par\medskip

\noindent\textbf{Crop growth (24 entries).} Experimental Crop Name(s); Seed Pre-treatment; Genotype/Cultivar; Trait; Photosynthetic Rate (Pn); Respiration Rate; Stomatal Conductance (gs); Transpiration Rate (Tr); Chlorophyll Content; Intrinsic Water Use Efficiency (Pn/gs; labeled WUE in the extraction schema); Specific Leaf Area (SLA); Leaf Area; Plant Height; Root Length Density (RLD); Root Length; Root Surface Area; Root Volume; Root Diameter; Proline, Soluble Sugar Content; Specific Root Length (SRL); Root:Shoot Ratio, Root:Stem:Leaf Ratio; Crop Water Use Efficiency (CWUE); Crop Nutrient Content; Crop Salt Content.\par\medskip

\noindent\textbf{Post-treatment effects (37 entries).} Biomass; Yield; Saline-Alkali Land Type; Electrical Conductivity (EC); Total Soluble Salt Content; pH; pH Measurement Method; Exchangeable Sodium Percentage (ESP); Sodium Adsorption Ratio (SAR):; Cation Exchange Capacity (CEC); Sodium Ion (Na$^{+}$) Content; Calcium Ion (Ca$^{2+}$) Content; Magnesium Ion (Mg$^{2+}$) Content; Chloride Ion (Cl$^{-}$) Content; Sulfate Ion (SO$_{4}$$^{2-}$) Content; Bicarbonate (HCO$_{3}$$^{-}$) Content; Carbonate (CO$_{3}$$^{2-}$) Content; Soil Organic Matter (SOM) Content; Soil Texture; Soil Bulk Density; Total Porosity; Soil Temperature; Calcium Carbonate (CaCO$_{3}$) Content; Available Phosphorus (P) Content; Available Potassium (K) Content; Total Nitrogen (N) Content; Ammonium Nitrogen (NH$_{4}$$^{+}$-N) Content; Nitrate Nitrogen (NO$_{3}$$^{-}$-N) Content; Trace Element Content; Field Capacity (FC); Permanent Wilting Point (PWP); Soil Water Content (SWC); Saturated Hydraulic Conductivity (Ks); Capillary Rise Height; Groundwater Depth; Irrigation Water EC; Annual Evaporation/Precipitation Ratio.\par\medskip

\noindent\textbf{Comprehensive benefits (5 entries).} Crop Yield Increase Rate; Soil Salinity/Alkalinity Reduction Rate; Amendment Input-Output Ratio; Soil Organic Carbon (SOC) Increment; Microbial Quantity and Diversity.\par\medskip
\endgroup

\subsection{Planning Instruction and Evaluation Rubric}

The instruction in the training examples asks: ``Please suggest suitable reclamation measures for this saline-alkali land and explain the reclamation effects.'' The target is organized as Land Analysis, Reclamation Measures, and Reclamation Effects. At inference, a budget-aware request adds the user budget to the site description and asks for a technically feasible comprehensive plan with controlled costs.

Automatic plan assessment uses precision, expertise, practicality, and innovation. Judges compare both plans and provide an explanation before selecting a preference. Expert questionnaires use anonymized plans, a 0--10 scale, and space for written feedback. Repeated expert ratings of the same questionnaire are averaged for system comparison, while individual ratings are retained for ICC estimation.

\section{Illustrative Evidence-to-Plan Instance}
\label{app:example}

Table~\ref{tab:case} presents selected fields from a SA-Bench source case, linking its initial conditions with the reported treatment and responses over the study duration.

\begin{table}[H]
\caption{Selected fields from the Linze County evidence-to-plan example. Outcomes were reported for the source treatment and duration.\label{tab:case}}
\small
\begin{tabularx}{\textwidth}{P{3.2cm}L}
\toprule
\textbf{Field} & \textbf{Source-case content}\\
\midrule
Location and soil & Xiaotun Township, Linze County, Gansu, China; Heihe River Basin; meadow solonchak.\\
Initial conditions & pH 8.45; salt content 8.18 g\,kg$^{-1}$; groundwater depth 1.50 m; bulk density 1.43 g\,cm$^{-3}$; organic matter 8.50 g\,kg$^{-1}$.\\
Climate & Mean annual precipitation 109 mm; mean annual evaporation 1,966 mm; mean annual temperature 7.8 $^\circ$C.\\
Reported intervention & Three years of alfalfa cultivation with manure, mineral fertilizer, and irrigation. The full source instance specifies treatment quantities and timing.\\
Reported soil response & At 0--20 cm, pH decreases from 8.45 to 8.20; salt content decreases by 4.63 g\,kg$^{-1}$; reported desalination rate 56.60\%; bulk density decreases by 0.24 g\,cm$^{-3}$.\\
Subsequent yield & Spring wheat 6.57 t\,ha$^{-1}$ and maize 9.84 t\,ha$^{-1}$ after ploughing the alfalfa.\\
Evidence identity & \emph{Study on Saline Soil Resources in the Heihe River Basin and the Effects of Alfalfa on Soil Improvement, Fertility Enhancement, and Ecological Restoration} (title recorded in the evidence instance).\\
\bottomrule
\end{tabularx}
\end{table}

The source instance explains the intervention through plant cover, root effects on structure, organic matter inputs, and nutrient replenishment. The explanatory target preserves the reported study conditions and connects them with the intervention and its measured responses.

An illustrative SA-Bench MCQ describes the Yinbei area of the Ningxia Yellow River Irrigation District, with silty fine sand, poor permeability, shallow groundwater, and drainage-related salinization. The options are:
\begin{enumerate}[label=\Alph*.]
\item Subsurface pipe drainage combined with salt-tolerant plant cultivation.
\item Systematic improvement of drainage channels, including main ditches, branch ditches, and renovation of electric drainage stations.
\item Application of gypsum combined with deep plowing and soil turning.
\item Flood irrigation for salt leaching and straw mulching.
\end{enumerate}
The designated source-supported answer is B, illustrating how the intervention is matched to the recorded drainage-related cause and the corresponding source case.

\section{Category-Level Measure-Selection Results}
\label{app:categories}

Table~\ref{tab:categories} preserves the original category-level results. The overall denominator is 936 MCQs. Unassigned categories are excluded from the breakdown, and label coverage differs by dimension; category counts therefore should not be summed across dimensions or assumed to exhaust the overall set.

\begin{table}[htbp]
\caption{Measure-selection accuracy (\%) without retrieval. QwQ denotes QwQ-32B; Qwen3 denotes Qwen3-Max; V3 and R1 denote DeepSeek-V3 and DeepSeek-R1; Gemini denotes Gemini-2.5-Pro.\label{tab:categories}}
\small
\setlength{\tabcolsep}{3pt}
\begin{tabularx}{\textwidth}{llr*{7}{>{\centering\arraybackslash}X}}
\toprule
\textbf{Dimension} & \textbf{Category} & \textbf{$n$} & \textbf{Wenyao} & \textbf{QwQ} & \textbf{Qwen3} & \textbf{V3} & \textbf{R1} & \textbf{GPT-5} & \textbf{Gemini}\\
\midrule
Severity & Mild & 171 & \textbf{93.57} & 74.85 & 82.46 & 81.29 & 80.12 & 78.36 & 80.12\\
 & Moderate & 185 & \textbf{96.22} & 80.54 & 84.86 & 84.86 & 78.38 & 82.16 & 80.54\\
 & Severe & 108 & \textbf{95.37} & 81.48 & 84.26 & 86.11 & 81.48 & 76.85 & 86.11\\
\midrule
Soil & Chloride & 212 & \textbf{96.23} & 85.85 & 91.04 & 90.09 & 86.32 & 88.68 & 86.32\\
 & Sulphate & 48 & \textbf{93.75} & 70.83 & 81.25 & 81.25 & 75.00 & 75.00 & 83.33\\
 & Sodic & 285 & \textbf{96.14} & 80.70 & 83.86 & 84.56 & 78.95 & 81.05 & 82.46\\
\midrule
Land & Coastal & 273 & \textbf{95.97} & 86.81 & 91.21 & 90.11 & 85.71 & 88.28 & 86.08\\
 & Inland & 383 & \textbf{95.04} & 78.33 & 81.46 & 83.81 & 79.63 & 78.85 & 82.77\\
 & Secondary & 230 & \textbf{96.52} & 84.78 & 89.13 & 90.87 & 86.96 & 87.39 & 85.22\\
\midrule
Measure & Engineering & 334 & \textbf{96.71} & 93.71 & 95.51 & 95.21 & 93.71 & 95.81 & 91.92\\
 & Agronomic & 109 & \textbf{94.50} & 75.23 & 79.82 & 82.57 & 74.31 & 77.06 & 86.24\\
 & Chemical & 261 & \textbf{95.02} & 77.39 & 84.67 & 84.29 & 80.84 & 78.93 & 81.23\\
 & Biological & 202 & \textbf{96.53} & 76.24 & 78.71 & 83.66 & 76.24 & 74.75 & 77.23\\
\midrule
\multicolumn{2}{l}{Overall} & 936 & \textbf{95.83} & 83.01 & 86.86 & 87.93 & 83.97 & 84.19 & 84.72\\
\bottomrule
\end{tabularx}
\end{table}

\section{Additional Automatic Plan Comparisons}
\label{app:original_comparisons}

Table~\ref{tab:original_plan_comparisons} reports additional comparisons of Wenyao + RAG with the baseline systems. All rates use 69 region--budget comparisons per judge. The matched-evidence analysis appears in Table~\ref{tab:matched_rag}.

\begin{table}[htbp]
\caption{Preference rates (\%) favoring Wenyao + RAG across 69 region--budget comparisons. A dash denotes an omitted self-comparison.\label{tab:original_plan_comparisons}}
\small
\begin{tabularx}{\textwidth}{Lcc}
\toprule
\textbf{Comparison model} & \textbf{DeepSeek-R1 judge} & \textbf{GPT-5 judge}\\
\midrule
QwQ-32B & 79.71 & 76.81\\
Qwen3-Max & 76.81 & 75.36\\
DeepSeek-V3 & 63.77 & 60.87\\
GPT-5 & 59.42 & ---\\
Gemini-2.5-Pro & 60.87 & 75.36\\
DeepSeek-R1 & --- & 65.22\\
\bottomrule
\end{tabularx}
\end{table}

\clearpage
\bibliographystyle{unsrtnat}
\bibliography{references}
\end{document}